\documentclass{article}

\usepackage[preprint]{corl_2026} 

\title{Discriminative Barrier Functions for Safe Adversarial Imitation Learning from Observation}

\author{
  \makebox[\textwidth]{\hfill
    Anubhav Vishwakarma\footnotemark[1] \quad
    Bhaumik Mehta \quad
    Caleb Hsu \quad
    Byron Boots \quad
    Tyler Han$^{\dagger}$ \quad
    Karen Leung$^{\dagger}$
  \hfill}\\[4pt]
  \makebox[\textwidth]{\hfill University of Washington\hfill}
}
\usepackage{xcolor}
\usepackage{amsmath}
\usepackage{amssymb}
\usepackage{mathtools}
\usepackage{hyperref}
\usepackage{cleveref}
\usepackage{url}
\usepackage{float}
\usepackage{booktabs}
\usepackage{subcaption}
\usepackage{svg}
\usepackage{graphicx}
\usepackage{wrapfig}
\usepackage[export]{adjustbox}
  
\usepackage{amsmath, amssymb}
\usepackage[labelfont=bf]{caption}

\usepackage{colortbl}
\usepackage{enumitem}
\usepackage{algorithm}
\usepackage{algorithmic}
\usepackage{arydshln}
\usepackage{fancyvrb}
\usepackage{tablefootnote}
\newcommand{\safe}{\ensuremath{\mathcal{S}}}

\newcommand{\unsafe}{\ensuremath{\mathcal{U}}}

\newcommand{\ent}{\ensuremath{\mathbb{H}}}
\newcommand{\R}{\ensuremath{\mathbb{R}}}
\newcommand{\tran}{\ensuremath{\mathcal{T}}}

\DeclareMathOperator*{\argmax}{argmax}
\newcommand{\grayrow}{\rowcolor{gray!10}}
\newcommand{\adddbf}[1]{\textcolor{blue}{#1}}

\newcommand\blfootnote[1]{%
  \begingroup
  \renewcommand\thefootnote{}\footnote{#1}%
  \addtocounter{footnote}{-1}%
  \endgroup
}
\begin{document}
\maketitle
\blfootnote{\textsuperscript{*}Corresponding author: av72@uw.edu \quad \textsuperscript{$\dagger$}Equal advising.}

\begin{abstract}
Inverse Reinforcement Learning (IRL) algorithms are powerful tools for learning from and generalizing expert demonstrations, but they often rely on unconstrained exploration, rendering them unsafe for real-world deployment. Meanwhile, Control Barrier Functions (CBFs) can guarantee the safety of control systems, but the analytical design of CBFs can be time-consuming and esoteric. In this work, we address these limitations jointly by constraining reward function candidacy during IRL to the space of CBFs, yielding a formulation that exhibits safe online control with continuous experiential improvement. Crucially, this framework enables the data-driven recovery of barrier functions directly from unlabeled expert observations. We demonstrate that the recovered barrier function is robust to unsafe states entirely absent from the expert data. Furthermore, we benchmark our method against standard IRL baselines in a simulated navigation environment, demonstrating improved safety performance. Finally, we investigate the trade-offs of planning-based versus policy-based IRL methods across both simulation and a real world obstacle avoidance task.
\end{abstract}



\section{Introduction}
Inverse Reinforcement Learning (IRL) serves as a principled approach for extracting the intrinsic objectives behind expert behavior, enabling the synthesis of robust, generalizable imitation policies~\citep{ng_algorithms_2000, GAIL_similar_cite_1, max_entropy_irl, helicopter_irl_beyond_grid_world}. However, most IRL methods rely on action-labeled demonstrations, which are often unavailable or expensive to collect in real-world robotics~\citep{torabi_generative_2019}. In contrast, humans are capable of learning to imitate behaviors directly from visual observation without access to the demonstrator's precise actions, improving through experience; in robot learning, this setting can be referred to as Inverse Reinforcement Learning from Observation (IRLfO), in which a robot recovers a reward function and an imitative policy from state-only expert trajectories~\citep{torabi_generative_2019}. IRLfO methods typically learn online, refining the policy through exploration in the environment~\citep{torabi_generative_2019, han_model_2025}. 

However, exploration poses safety risks that limit the real-world deployment of these algorithms~\citep{JMLR:v16:garcia15a}. Humans, by contrast, exhibit the ability to avoid \textit{unsafe states} when learning from expert observations. In cognitive science, the suppression of actions at unsafe states is associated with inhibitory control~\citep{inhibit_control_0, inhibit_control_1, control_inhibition}. Yet existing IRLfO methods focus solely on behavior fidelity and provide no explicit mechanism for avoiding unsafe states during the exploration and learning process~\citep{torabi2018behavioralcloningobservation, torabi_generative_2019, han_model_2025}. To address this limitation, we study the following problem within the IRLfO setting: \textit{Given expert observational trajectories, how can a robot learn an imitative policy that also performs inhibitory control when encountering unsafe states during learning?}

To instantiate this notion of inhibitory control, we look to control theory, in particular, Control Barrier Functions (CBFs)~\citep{ames_control_2019}. CBF theory consists of a barrier function and a dynamics model, and, under certain conditions, it can render a set control-invariant (i.e., there exists a control that ensures the system remains inside the (safe) set indefinitely). Owing to these guarantees, CBFs have been widely employed as explicit safety layers in applications such as adaptive cruise control~\citep{CBFs_cruise_control}, aerial robotics~\citep{CBFs_aerial_robots}, and legged locomotion~\citep{CBFs_legged_locomotion}. However, manually designing task-specific barrier functions to form CBFs for safety-critical systems remains challenging, analogous to reward engineering in reinforcement learning (RL) ~\citep{cbfs_difficult, cbfs_difficult_2, cbfs_difficult_3}.
\begin{figure}[H]
    \centering
    \includegraphics[width=\textwidth]{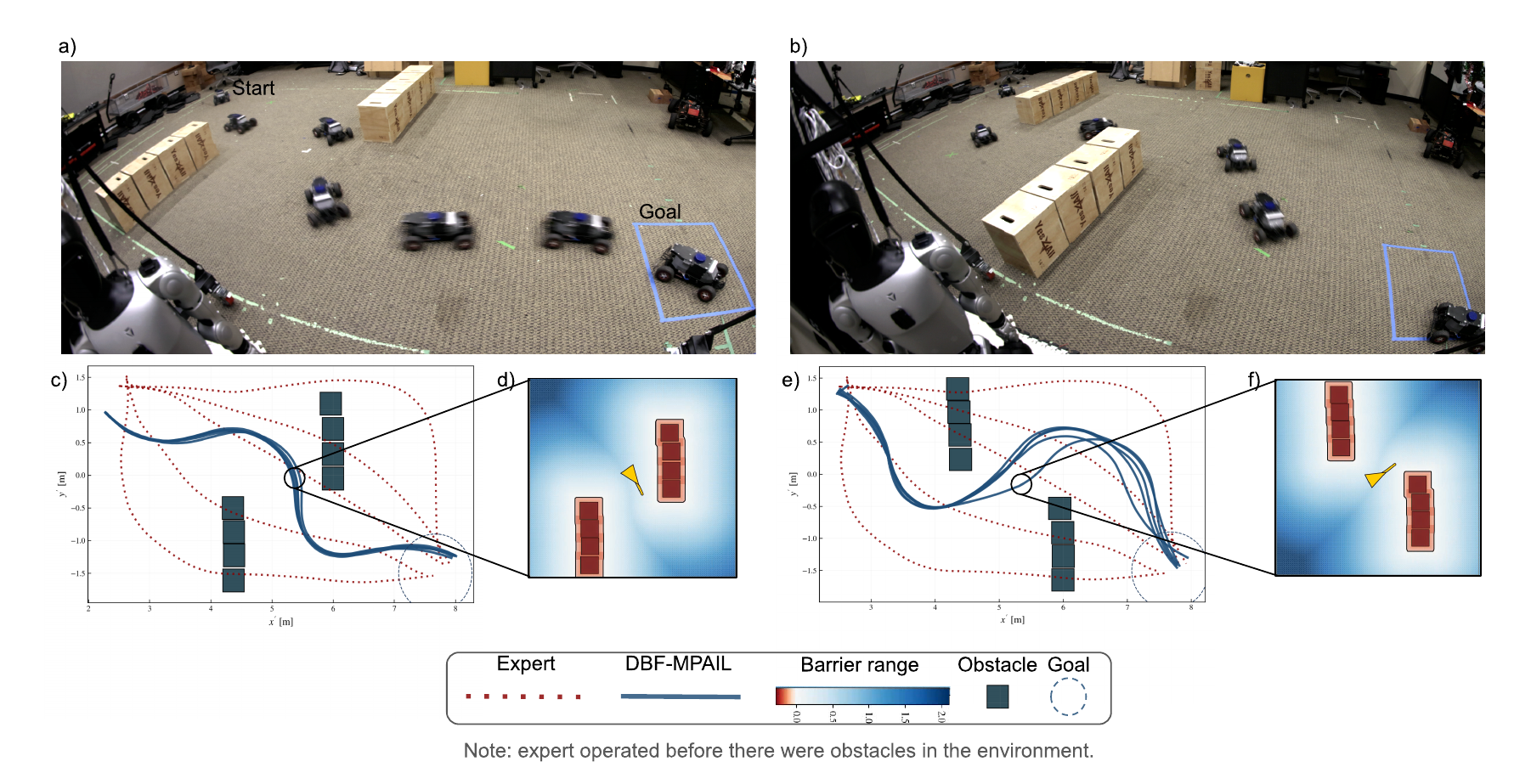}
    \caption{Hardware deployment across two obstacle configurations unseen during training (a,b), demonstrating generalization to unseen unsafe observations. Robots are shown mid-inference navigating toward the goal. (c,e) Closed-loop trajectories for DBF based planning method with expert reference (dotted). (d,f) bird's-eye view map of the robot representing the barrier function range for the map region. See \Cref{fig:real-sim-real} and \Cref{tab:hardware_results} for performance on additional unseen scenarios. \Cref{fig:baselines_plots} shows baseline rollouts in above two scenarios.}
    \label{fig:Hardware_validation}
\end{figure}
In this work, we investigate the hypothesis that constraining reward function candidacy with the forward invariance property required by CBFs may lead to the recovery of both the barrier function and a policy that learns to inhibit its control near unsafe observations through experience. In particular, we employ a discrete form of CBFs to relax the continuity and differentiability requirements of the barrier function during training.
\textit{In this work, we propose DBFs (Discriminative Barrier Functions) for IRLfO, which recovers a learned barrier function and an imitative policy jointly from state-only expert observations, enabling safe online imitation learning without action supervision and hand-designed constraints.}
We summarize our key contributions as follows:

\textbf{Learning Interpretable Barrier Functions:} We demonstrate that restricting the cost function class within the IRLfO framework to Discrete-time CBFs~\citep{DBFs} yields highly interpretable and generalizable barrier functions; this constrained objective to synthesize cost function through IRLfO is called DBF. These functions successfully capture expert intent while inherently discriminating against unsafe observations, such as obstacles completely absent from the expert demonstrations.

\textbf{Safe Online Learning from Observation:} We show that incorporating DBF regularization into the reward function search significantly reduces collisions in online navigation settings. This mitigates the risks of environment interaction during IRLfO, paving the way toward safe online deployment in the real world.

\textbf{Empirical Validation of the DBF Formulation:} We test and compare DBF across both policy-based and planning-based IRLfO frameworks, demonstrating its efficacy in obstacle avoidance tasks through both simulation and real-world robotic experiments unseen during training time, thereby showcasing the generalization capability of DBF.%


\section{Related Work}
In this section, we review two related lines of work: methods that learn barrier functions from data, and safe reinforcement and imitation learning methods that recover safe policies. Both connect to our goal of learning inhibitory control online from unlabeled observations.

\subsection{Learning Control Barrier Functions}
CBFs can be challenging to design by hand for general, high-dimensional systems, yet their safety desiderata are intuitive: forward invariance within the safe set and recovery enforcement when violated \citep{ames_control_2019}. These properties make CBFs an attractive structural object for robot learning, where they have already garnered substantial research attention \citep{ames_control_2019, guerrier_learning_2024, neural_cbf}. A common approach to learning CBFs from data exploits discriminative boundaries between safe and unsafe regions ~\citep{srinivasan_synthesis_2020, robey_learning_2020}. For instance, \citet{srinivasan_synthesis_2020} optimizes for a separating hyperplane in a kernel-induced feature space. Along a related line, \citet{yang_safe_2023, yang_enhancing_2024} employ Adversarial Inverse Reinforcement Learning (AIRL) as a pre-training step to obtain a policy, which is subsequently used to recover a CBF. More recently, ~\citet{paper:nakamura-cbf-2026} learn a value-function-based CBF from action-labeled demonstrations, similarly requiring a pre-specified safe set and offline supervision ~\citep{neural_cbf}. In contrast, a parallel line of work learns barrier-like certificates directly from demonstration data without explicit safe / unsafe labels ~\citep{i-DBF, robey_learning_2020}, or imposes conservatism toward out-of-distribution states ~\citep{Tabbara_ccbf_2025, v_OCBF}. However, these methods uniformly require action-labeled demonstrations, operate strictly offline, and rely on a pre-specified unsafe set, limiting their applicability to settings where safety structure must be inferred from unlabeled observations alone. Our work addresses this gap by learning a discrete barrier function directly within the adversarial imitation loop, requiring neither action labels, safety labels, nor offline pretraining.%

\subsection{Safe Reinforcement and Imitation Learning}
Towards high-dimensional systems that learn from interaction, safe RL methods aim to adhere to safety-oriented constraints while simultaneously maximizing reward \citep{gu_review_2024}. These approaches naturally depend on a pre-specified constraint set and do not address settings in which the safety structure must be derived from data. SafeDICE \citep{safe_dice} addresses offline safe imitation learning by correcting stationary distributions using labeled non-preferred demonstrations, avoiding constraint-violating behavior without explicit reward learning. However, it operates purely offline, requires action-labeled demonstrations and explicit non-preferred labels, and lacks any mechanism for online adaptation, assumptions, and limitations that our framework removes entirely. Closer to our method, CBF-RL \citep{cbfrl} incorporates a hand-specified CBF as a constraint term in the policy loss during RL training, internalizing safety without requiring a runtime safety filter at inference. However, the CBF itself must be provided a priori and is not learned from data. While our DBF cost plays an analogous role, discouraging actions that push toward unsafe regions, it is  learned entirely from unlabeled observations rather than  derived from a hand-specified model, and operates within an adversarial imitation loop rather than a pure RL setting.%


\section{Discriminative Barrier Functions}
This section describes our methodology for learning a barrier function through the Adversarial Imitation Learning (AIL) framework.%
\subsection{Preliminaries:}
Consider a Partially Observable Markov Decision Process (POMDP). In an unknown world state $s_w\in S_w$, the agent makes an observation $o\sim p(o|s_w)$.
From a history of observations $\mathbf{o}_{0:t}$, the agent perceives its \textit{state} $s_t\sim p(s|\mathbf{o}_{0:t})$ and performs \textit{action} $a \in \mathcal{A}$. Through a predictive model $f:S \times \mathcal{A} \rightarrow S$, the agent self-predicts forward in time for a model-based method.
Partial observations of $S$ are desirably used for demonstrations to perform IRL and AIL, as full observation history would quickly become intractable~\citep{risk_aware_cost_maps}. The expert $\pi_E$ may be acting to optimize some reward based on current and next state $r_w(s,s')$ in the POMDP, which may also potentially encode safety constraints. While in this work we do not assume its availability, known constraints can be employed seamlessly as we discuss in \Cref{sec:abf}.
\subsection{Inverse Reinforcement Learning from Observation}
While there are many IRL formulations from which to begin, ~\citet{ho_generative_2016} show how apprenticeship learning with regularization unifies various perspectives on IRL. In short, this objective seeks cost functions $c(\cdot)$ under which the performance of the expert is maximized while other \textit{optimal} policy (i.e. learner, $\pi$) candidates are minimized:
\begin{equation}
    \text{IRLfO}_{\psi}(\pi_E) = \argmax_{c \in \mathbb{R}^{S \times S}} -\psi(c) + \underbrace{\min_{\pi \in \Pi} -\lambda \ent(\pi) + \mathbb{E}_{\pi}[c(s, s')]}_\textbf{Entropy-regularized Policy Optimization}
 - \mathbb{E}_{\pi_E}[c(s, s')],\label{eq:irlfo}
\end{equation}
where the observation-only cost $c(s,s')$ \citep{torabi_generative_2019} is introduced, reflecting the absence of expert action data in our problem setting.

It is through the cost regularizer, $\psi:\mathbb{R}^{S\times S}\rightarrow \mathbb{R}$, that \citet{ho_generative_2016} instantiate other IRL algorithms. Likewise, to introduce our specific form of regularization for our DBF formulation, a brief introduction to CBFs is required.

\subsection{Discrete-Time Control Barrier Functions}
Let safe states be the set $\safe$ and unsafe states $\unsafe$.
Towards barrier functions, we also introduce the \textit{specification function} $h(\cdot)$ such that, $\safe\coloneqq\{s\in\mathcal{S}~\vert~h(s) \geq 0\}$.
Finally, $h$ can be certified as a Discrete-Time CBF if
\begin{equation}
    \sup_{a\in\mathcal{A}}h(f(s,a)) - h(s) \geq -\alpha(h(s)),\label{eq:dcbf}
\end{equation}
where $\alpha$ is a class $\mathcal{K}$ function. Under this condition, CBFs guarantee that there is an action which maintains forward invariance and stability. For instance: if $h(s_t) \rightarrow 0^+$ (i.e. agent approaches unsafe from safe states), the allowable rate at which the agent may continue approaching $\unsafe$ goes to 0 thereby stably avoiding unsafe states; if $h(s_t) < 0$ (i.e. $s_t \in \mathcal{U}$ ), then there is an action for which the agent can increase $h$ for $s_{t+1}$ thereby returning to safety.

\subsection{Discriminative Barrier Functions via Cost Regularization}\label{sec:dbf-derivation}
Now equipped with the IRL optimization objective and CBF safety constraints, we can proceed with the Discriminative Barrier Function (DBF) formulation.
First, we observe a simple relationship between AIL and CBFs, which will allow us to view CBFs as a constraint on the AIL objective.

\begin{wrapfigure}{r}{0.5\columnwidth}
    \vspace{-5pt}
    \centering
    \includegraphics[width=0.5\columnwidth]{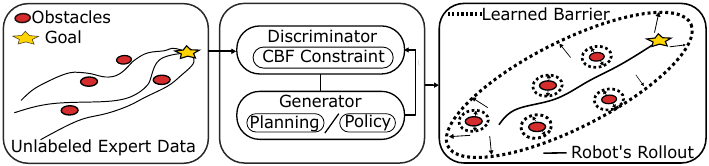}
    \caption{Overview of DBF: a barrier function and imitation policy are recovered jointly from unlabeled expert observations within an AIL framework. The discriminator enforces CBF constraints, classifying expert states as safe and learner rollouts as unsafe, while the generator (planning or policy) optimizes against the learned barrier.}
    \label{fig:method}
    \vspace{-10pt}
\end{wrapfigure}

In AIL, an ideal discriminator $D^*$ achieves a boundary which perfectly separates expert states $s\in\mathcal{D}_E$ from learner states $s\in\mathcal{D}_\pi$. Meanwhile for CBFs, the barrier function $h$ can be viewed as separating safe states $s\in\safe$ from unsafe states $s\in\unsafe$ as illustrated in ~\Cref{fig:method}. 
These binary classification objectives appear practically adjacent: \textit{can a barrier function act as a discriminator?} Rather than learning only to discriminate expert and learner behavior, we also seek to discriminate ``safe'' from ``unsafe'' behavior.  \textit{But, using only positive expert demonstrations, how can we define safety meaningfully?}
By adopting a control-theoretic definition of safety, we can constrain the discriminator to classify a transition $(s,s')$ based on the condition in \Cref{eq:dcbf}, through the \textbf{Dynamic Constraint Lagrangian} $q(s,s')$:
\begin{equation}
    q_h(s,s') \coloneqq h(s') - h(s) + \alpha(h(s))
\end{equation}
which tends more positive for safer transitions and conversely more negative for less safe transitions.

Returning our attention to the IRL objective in \Cref{eq:irlfo}, consider restricting the cost function space to $\mathbb{R}^S$ rather than in unconstrained IRL which searches all of $\mathbb{R}^{S\times S}$.
\begin{equation}
    \mathcal{C}_\text{CBF} \coloneqq \left\{ c~|~c(s,s') = q_h(s,s'),~h\in\R^S\right.
    \mathrm{s.t.} \quad h(s)>0,~\forall s\in\safe
    \left.\quad h(s) < 0,~\forall s\in\unsafe\right\}.
\end{equation}
That is, we wish to search through cost function candidates that are \textit{also Control Barrier Functions} (thereby also imposing control-theoretic structure on the cost function). By combining this restriction on $c$ with the convex regularization from ~\citet{ho_generative_2016}, we derive (see \Cref{appendix:dbfs-derivation} for proof) the \textit{DBF objective}, 
\begin{align}
\label{eq:dbf}
    \arg\max_{h \in \mathbb{R}^S} \quad &
    \underbrace{
        \mathbb{E}_{\tau_{\text{safe}}}\!\left[D_h(s,s')\right]
        -
        \mathbb{E}_{\tau_{\text{unsafe}}}\!\left[D_h(s,s')\right]
        }_{\mathcal{L}_{\textbf{Wasserstein-GAN}}}
    -
    \underbrace{
        \mathbb{E}_{\mathcal{\hat{T}}}
        \left[
        \left(
        \left\|\nabla_{(s, s')} D_h(s, s')\right\|_2 - 1
        \right)^2
        \right]}_{\mathcal{L}_{\textbf{Gradient Penalty}}}
    \\
    & \qquad
        \text{s.t.} \quad
        \begin{cases}
            h(s) > 0, & \forall s \in \mathcal{S}_{\text{safe}}
            \\
            h(s) < 0, & \forall s \in \mathcal{S}_{\text{unsafe}}
        \end{cases}
\end{align}
%
where $D_h(s,s') = q_h(s,s')$, $\hat{\tran} \sim \eta \tran_{safe} + (1-\eta) \tran_{unsafe}$ and $\sigma(x) \coloneqq \left(1 + e^{-x}\right)^{-1}$ is the sigmoid function. Note that for the barrier function, to achieve stable training of GAN's we use a Wasserstein GAN formulation, which encourage 1-Lipschitz continuity via a gradient penalty \citep{WGAN, wgan_gp}.

\subsection{Learning DBFs via Adversarial Imitation Learning}\label{sec:abf}
What remains is how this objective should be computed.
Since we have not assumed that safety constraints are known, we make the simplifying assumptions that (i) learner data is unsafe, $\tran_\text{unsafe}\sim\pi$ and (ii) expert data is safe, $\tran_\text{safe}\sim\pi_E$.
Meaning, we can then employ Adversarial Imitation Learning to compute the expectations in \Cref{eq:dbf}. Altogether, the AIL objective with DBFs can be summarized:
\begin{align}
\label{eq:loss_func}
    \min_{\pi \in \Pi} \max_{h \in \mathbb{R}^S} \quad
    &  \nonumber
    \lambda_{\text{wgan}} \Big[
        \mathbb{E}_{\pi_E}[D_h(s, s')] 
        -
        \mathbb{E}_{\pi}[D_h(s, s')]
        \Big]
    - 
    \lambda_{\text{GP}} \,
        \mathbb{E}_{\hat{\pi}}
        \Big[
        (\|\nabla_{(s, s')} D_{h}(s, s')\|_2 - 1)^2
        \Big] \\
    \quad & - \ent(\pi) + \lambda_{\text{sign}} \Big[
    \underbrace{
        \mathbb{E}_{s \sim \pi_E} \big[(\delta + h(s))_+\big]
        +
        \mathbb{E}_{s \sim \pi} \big[(\delta - h(s))_+\big]
        }_{\textbf{hinge loss}}
    \Big] 
\end{align}
This formulation optimizes the DBF directly from data without requiring explicit, pre-defined safety labels, leveraging the adversarial framework to adaptively separate safe and unsafe states online.

In the setting where safety constraints \textit{are known}, it is straightforward to incorporate this additional information, for example, by also populating $\tran_\text{safe}$ conditionally from the learner or by decomposing $h(s) = \min\{\bar{h}(s), h_\theta(s)\}$ where $\bar h$ and $h_\theta$ are prior and learned barrier functions, respectively. In addition, all demonstrations need not be safe as assumed by $\tran_\text{safe} \sim \pi_E$. It is straightforward to include demonstrations that are unsafe directly in $\tran_\text{unsafe}$. While we do not explore these extensions here due to their assumption of additional safety knowledge, they remain as promising future work in more applied settings.

As in ~\citet{robey_learning_2020}, our intuition behind states being drawn from closed and non-empty safety sets requires notions of $\epsilon$-nets. Put simply, each demonstrated state subsumes an $\epsilon$ ball of states around it, implying that the expert operates with some minimum distance to the unsafe boundary. This is equivalent to constraining the value of the barrier function at safe and unsafe states while limiting the Lipschitz constant of the function, which (along with gradient penalties) is coincidentally common practice for stabilizing AIL algorithms ~\citep{difficult_distractor}.


	

\section{Experimental Results}
\label{sec:result}
\begin{algorithm*}[tb]
\caption{Discriminative Barrier Function based Adversarial Inverse Reinforcement Learning}
\label{alg:inverse_rl}
\begin{algorithmic}[1]
    \STATE Obtain state-only expert observation trajectories $\tau_E$
    \STATE Initialize policy $\pi_\phi$ and discriminator $D_{\adddbf{h_\theta}}$.
    \STATE Initialize experience replay buffer \adddbf{$\mathcal{B} \leftarrow \{\}$.}
    \FOR{step $t$ in $\{1, \dots, N\}$}
        \STATE Collect trajectories $\tau_i = (s_0, \dots, s_T)$ by executing $\pi_\phi$ and \adddbf{ $\tau_i \rightarrow\mathcal{B}$}.
        \STATE Sample interpolations \adddbf{$\hat{\tau} \sim \eta \tau_{E} + (1-\eta) \tau_{u}$}; where \adddbf{$\tau_{u} \sim \mathcal{B}$}.
        \STATE Update discriminator parameters $\theta$ by:
        \adddbf{\begin{align*}
            \mathcal{L}(\theta) = \;& \lambda_{\text{wgan}} \Big[ \mathbb{E}_{\tau_u}[D_{h_\theta}(s, s')] - \mathbb{E}_{\tau_E}[D_{h_\theta}(s, s')]\Big] + \lambda_{\text{GP}} \mathbb{E}_{\hat{\tau}} \Big[(\|\nabla_{(s, s')} D_{h_\theta}(s, s')\|_2 - 1)^2\Big] \\
            & - \mathbb{E}_{s \sim \tau_E} \big[(\delta + h_\theta(s))_+\big] - \mathbb{E}_{s \sim \tau_u} \big[(\delta - h_\theta(s))_+\big]
        \end{align*}}
        \STATE $r_\theta(s, s') \leftarrow \text{AIL}(D_{h_\theta}(s, s'))$ \Cref{tab:baselines}
        \STATE Update policy parameters $\phi$ using $r_\theta(s,s')$ with any policy optimization method.
    \ENDFOR
\end{algorithmic}
\end{algorithm*}
\begin{wraptable}{r}{0.40\columnwidth}
\centering
\scalebox{0.7}{
\begin{tabular}{lrr}
\toprule
    Algorithm & Reward & Learner \\
    \midrule
    \grayrow
    GAIL  & $\log D$              & PPO (Policy)   \\
    AIRL    & $\sigma^{-1} \circ D$ & PPO (Policy)   \\
    \grayrow
    MPAIL     & $\sigma^{-1} \circ D$ & MPPI (Planner) \\
    \bottomrule
    \end{tabular}
    }
    \caption{AIL Algorithm Summaries}
    \label{tab:baselines}
\end{wraptable}
We evaluate the DBF objective with different AIL \Cref{alg:inverse_rl}, including GAIL~\cite{ho_generative_2016}, AIRL~\cite{fu_learning_2018}, and MPAIL~\cite{han_model_2025}, summarized in \Cref{tab:baselines}. These algorithms are primarily differentiated by their reward structure and whether they are policy-based or planning-based. GAIL and AIRL are model-free methods, while MPAIL is a model-based AIL method that achieves strong out-of-distribution recovery through approximated policy value-function bootstrapping at the terminal state~\citep{han_model_2025}. 
In experiments we aim to answer the following questions:
\begin{enumerate}[label=\textbf{Q\arabic*}]
    \item [\textbf{Q1}] Can DBFs learn barrier functions that generalize to novel environments from unlabeled observations unseen during training? \label{q1}
    \item [\textbf{Q2}] How do DBF-augmented AIL algorithms compare against existing AIL baselines in terms of safety? \label{q2}
    \item [\textbf{Q3}] Can DBFs improve safety in real-world and safety-critical environments? \label{q3}
\end{enumerate}
\noindent\textit{Note:} In all experiments, the expected value of the discriminator in \Cref{eq:loss_func} is calculated off-policy.
\subsection{Navigation In Simulated Environment}\label{sec:env_setup}
\begin{figure}[h]
    \centering
    \includegraphics[width=\textwidth]{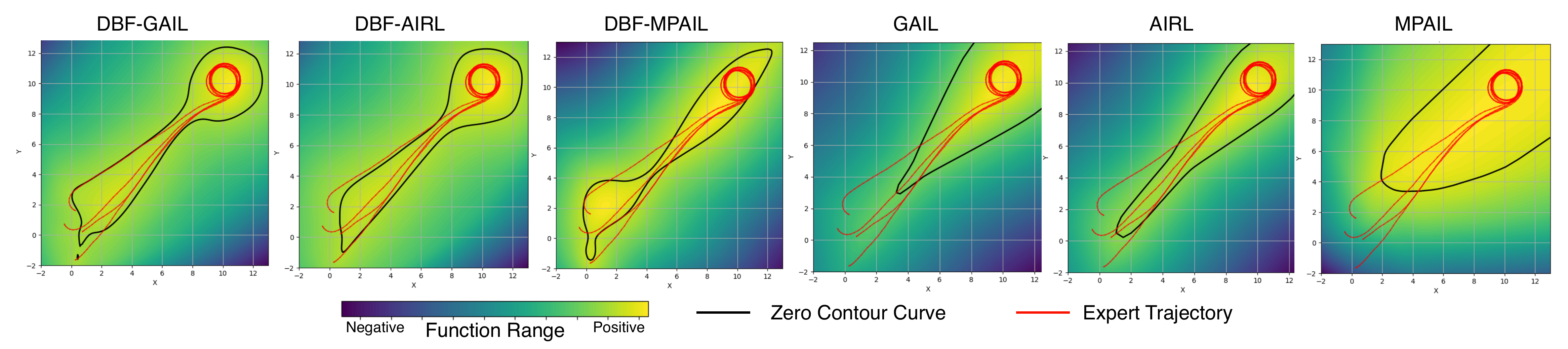}
    \caption{Qualitative comparison of the learned barrier function $h(s)$ for DBF-based methods and the reward function $r(s)$ for non-DBF variants of GAIL, AIRL, and MPAIL. For DBF methods, the learned barrier function $h(s)$ is plotted; for non-DBF baselines, the learned reward function $r(s)$ is plotted. The black curve denotes the zero contour, and the red curve denotes the expert trajectory. Here $s \in \mathbb{R}^2$ represents the agent's position.}
    \label{fig:no-obstacle-sim-bf-comparison}
    \vspace{-10pt}
\end{figure}

We first qualitatively analyze the learned boundary around expert trajectory in a simple no-obstacle simulation setting, where unsafe states can be defined as those not present in the expert data. We use MUSHR~\citep{srinivasa2019mushr}, a 10-DOF ego vehicle, as a testbed in the Isaac Lab \citep{isaaclab} simulation environment for navigation to a goal point.
\Cref{fig:no-obstacle-sim-bf-comparison} compares the barrier function learned by DBF-based methods, $h(s)$, with the single-state (position-only) reward function, $r(s)$, used by baseline AIL methods for learning the discriminative boundary around the expert trajectory. This comparison is necessary because, in standard AIL, the reward function may implicitly encode safety constraints. We retain the circling motion near the goal in the expert trajectory for qualitative comparison, since this behavior is more difficult to imitate from state-only observations, as corroborated by~\citep{difficult_distractor}. The zero contour curve serves as the discriminative boundary --- states inside are classified as safe ($h(s), r(s) > 0$) and states outside as unsafe ($h(s), r(s) < 0$). AIL algorithms equipped with DBFs generalize substantially better around the expert demonstrations, while also correctly modeling the challenging circling behavior at the goal. In contrast, the baselines collapse and overestimate the discriminative boundary near goal position.
\subsection{Q1: Barrier Function Robustness to Unsafe States not Present in Expert Data} \label{sec:ex_sub}
\begin{figure}[t]
    \includegraphics[width=\textwidth]{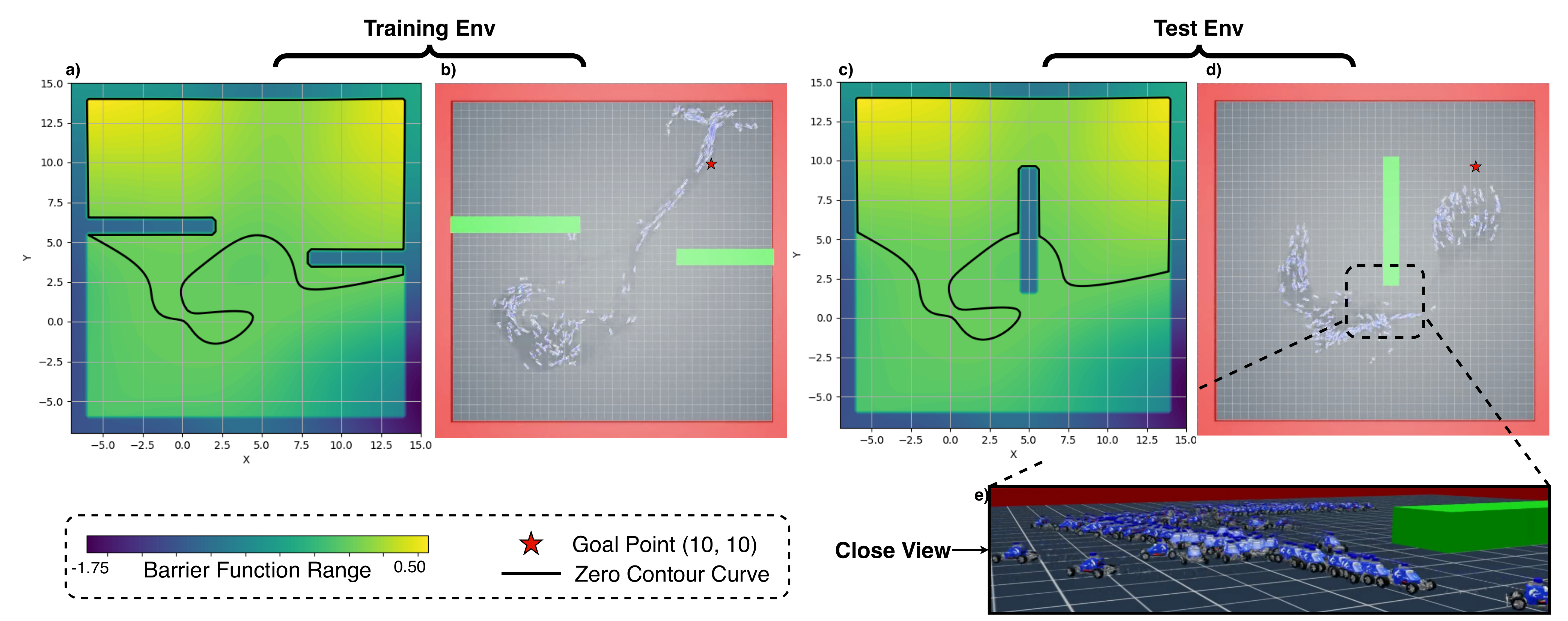}
    \caption{\textbf{Qualitative analysis of barrier function generalization synthesized via DBF-MPAIL.} The barrier function learned from unlabeled expert observations generalizes zero-shot (zero collisions) to a novel obstacle configuration unseen during training, correctly classifying the new obstacle as unsafe without retraining. \textbf{(a)} Learned continuous barrier function $h(s)$ from expert state-only observations (\Cref{fig:expert_trajectory}); warm colors indicate safe regions ($h(s) > 0$), cool colors unsafe regions ($h(s) < 0$). \textbf{(b)} Top view of the training environment with the marked goal point. \textbf{(c)} The same barrier function adapted to a novel test environment with a previously unseen vertical obstacle — note the unsafe region correctly emerging around the new obstacle. \textbf{(d)} Top view of the test environment showing agent trajectories navigating around the unseen obstacle. \textbf{(e)} Close-up view of robots inhibiting their control inputs to avoid obstacles in the unseen environment.}
    \label{fig:Q1_figure}
\end{figure}

\textbf{Experiment Setup:}
For this study, we use a maze environment enclosing the expert trajectories (\Cref{fig:expert_trajectory}), with horizontal wall positions randomized during training (\Cref{fig:Q1_figure}b) to promote generalization, the discriminator is evaluated with expectations from an off-policy replay buffer, exposing it to a diverse set of states beyond the current policy's distribution. A vertical wall is placed at the center of the environment during test time to investigate generalization to unseen obstacle configurations (\Cref{fig:Q1_figure}d). All experiments use linear $\mathcal{K}$-class functions unless otherwise specified.

\textbf{Qualitative Analysis:} We investigate robustness to unsafe states of the DBF-based method unseen in expert data after having established that DBF-based methods learn expressive discriminative boundaries (\Cref{fig:no-obstacle-sim-bf-comparison}), especially DBF-MPAIL, among having the highest variation in expressivity within the safe set, i.e., zero contour. We next analyze DBF-MPAIL in the obstacle-avoidance task using unlabeled expert observational trajectories to train the method. During training, the expert data is classified as safe; however, when learning to imitate in the simulation environment, the robot may encounter obstacles unseen in the expert data that hinder its navigation to the goal due to walls. As shown in \Cref{fig:Q1_figure} parts a and c, DBF-MPAIL preserves meaningful notions of safety even under these distribution shifts both while training and during evaluation, respectively. Although portions of the expert trajectories intersect obstacle regions, the learned barrier function adapts by rejecting unsafe transitions while maintaining connectivity to the demonstrated safe set.
\begin{wraptable}{r}{0.6\textwidth}
\vspace{-\intextsep}
\centering
\caption{Obstacle navigation evaluation (mean $\pm$ std across $N=5$ training seeds: 1, 2, 3, 22, and 312) in environment shown in \Cref{fig:Q1_figure}d.}
\label{tab:obstacle_eval}
\scalebox{0.85}{%
\begin{tabular}{lcc}
\toprule
Algorithm & Collision (\%) $\downarrow$ & Avg.\ Env Reward $\uparrow$ \\
\midrule
\grayrow
DBF-AIRL & $\mathbf{0.0 \pm 0.0}$ & $\mathbf{0.264 \pm 0.078}$ \\
DBF-GAIL & $\mathbf{0.0 \pm 0.0}$ & $\mathbf{0.300 \pm 0.014}$ \\
\grayrow
DBF-MPAIL & $\mathbf{0.0 \pm 0.0}$ & $\mathbf{0.338 \pm 0.022}$ \\
AIRL & $0.0 \pm 0.0$$^{*}$ & $0.246 \pm 0.041$ \\
\grayrow
GAIL & $20.5 \pm 25.4$ & $-0.030 \pm 0.174$ \\
MPAIL & $12.3 \pm 1.8$ & $0.455 \pm 0.029$ \\
\bottomrule
\end{tabular}%
}
\smallskip
\parbox{0.58\textwidth}{\small $^{*}$AIRL achieves zero collisions due to its reward structure, not explicit safety constraint.}
\vspace{-\intextsep}
\end{wraptable}
The continuity imposed by the barrier objective (\Cref{eq:loss_func}) forces the learned safe set to remain topologically consistent around the obstacle. 

\textbf{Quantitative comparison:} To quantify above analysis for all the DBF based AIL methods, we evaluate the performance of them in the test environment (\Cref{fig:Q1_figure}d) by comparing the total number of collisions out of the total number of robot spawns, as well as the average environmental reward, to quantify the task objective of reaching the goal while avoiding obstacles. The results are shown in \Cref{tab:obstacle_eval}.\footnote{In all experiments involving baselines, the reward structure was $r(s, s')$, whereas in \Cref{sec:env_setup}, $r(s)$ was used to qualitatively distinguish each method.} 
It is evident from the performance table that DBF objective strongly influences the performance of planning-based method, since the reward function (i.e., the CBF constraint) is directly used as a stage cost of MPPI planner, providing strong guidance toward safe actions. In addition, value bootstrapping at terminal states encodes safety context, providing guidance analogous to the role of the learned policy at terminal states leading to control inhibition, as shown in \Cref{fig:Q1_figure}e. Moreover, we observe improved performance in policy-based AIL methods: GAIL with the DBF formulation leads to zero collisions than its counterpart, indicating that the learned policy derived from this discriminative reward function does promote inhibitory control.
These comparisons demonstrate that the DBF objective learns a generalizable barrier function, as shown qualitatively in \Cref{fig:Q1_figure} and demonstrated performance quantitatively in \Cref{tab:obstacle_eval}.
\subsection{Q2: Safe Online Imitation Learning Performance Comparison and Safety Analysis} \label{sec:ex_sub}

For online performance evaluation we compare both imitation performance via normalized reward and safety via cost rate~\citep{Ray.et.al}: a metric that quantifies constraint-violation regret throughout learning in the same environment as Q1.

\begin{wrapfigure}{r}{0.5\columnwidth}
    \vspace{-15pt}
    \centering
    \includegraphics[width=\linewidth]{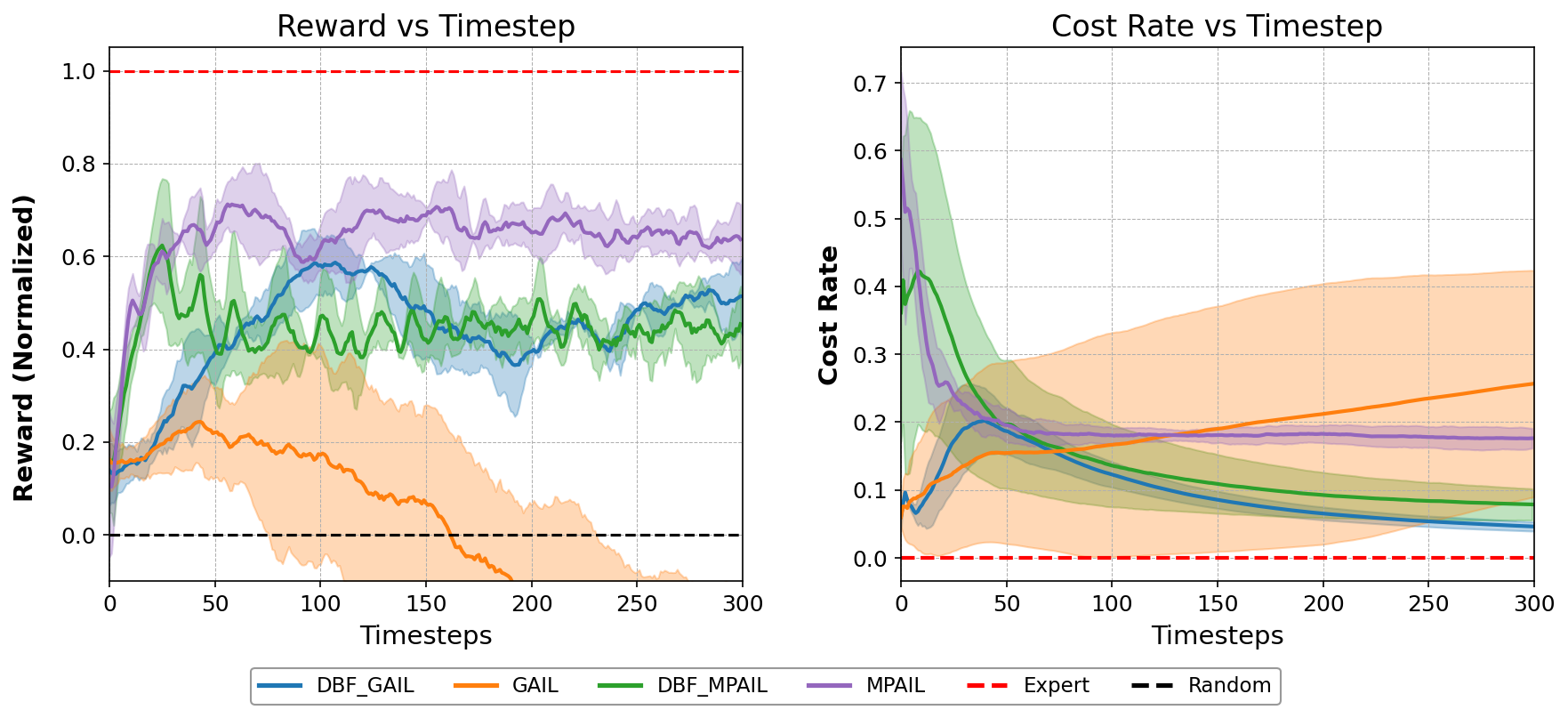}
    \caption{Normalized reward and cost rate during online learning, comparing DBF and non-DBF variants across all methods. Shaded regions represent standard deviation across training seeds. DBF-based methods consistently reduce cost rate while maintaining competitive reward.}
    \label{fig:benchmark-perf}
    
    \vspace{5pt}
    \includegraphics[width=\linewidth]{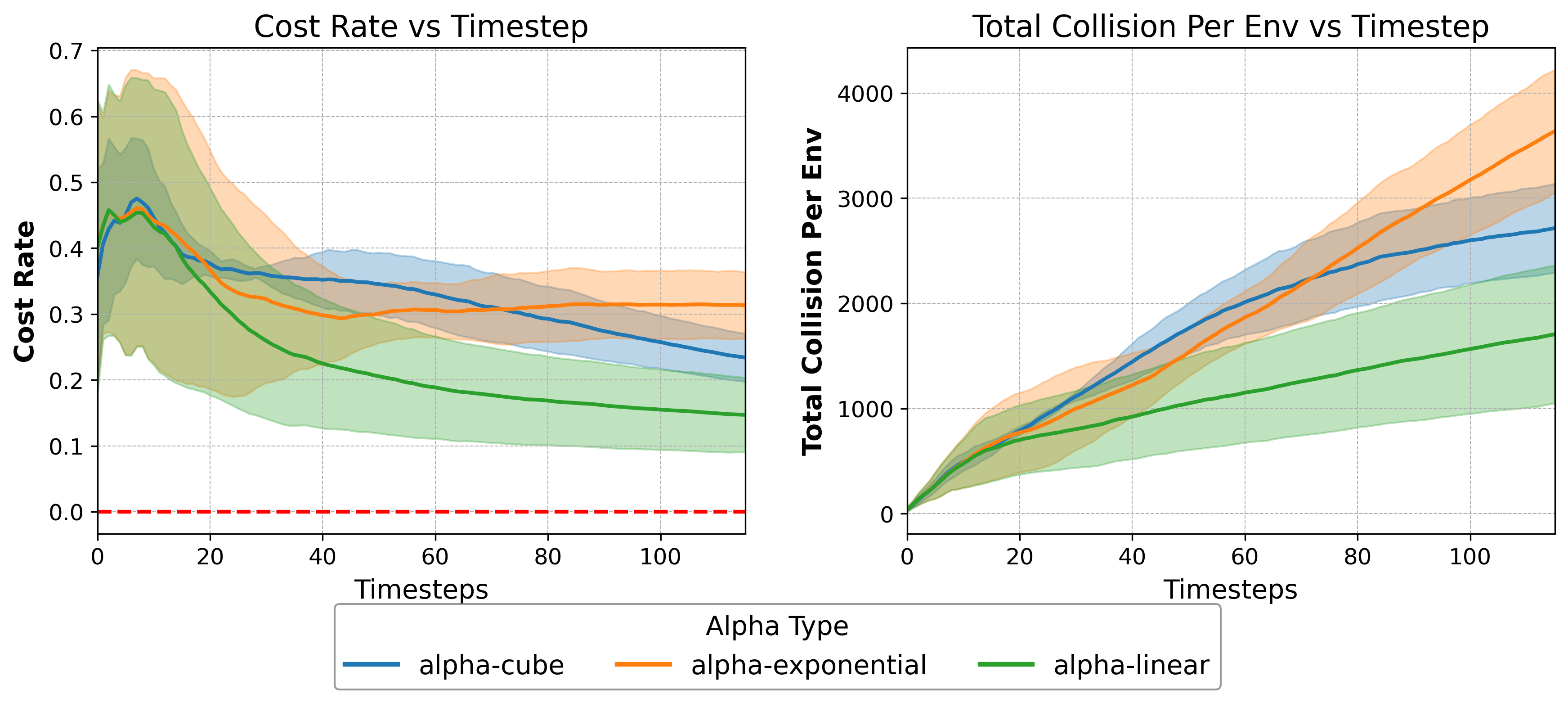}
    \caption{Effect of $\mathcal{K}$-class functions on cost rate and total collisions, evaluated on DBF-MPAIL. Linear $\mathcal{K}$-class functions achieve the fastest reduction in cost rate, consistent with their steeper slope in the discriminator logit range.}
    \label{fig:K-class}
    
    
    \vspace{-30pt}
\end{wrapfigure}

As shown in \Cref{fig:benchmark-perf}, DBF-based methods consistently reduce cost rate as learning progresses, while GAIL collapses entirely, failing to imitate the expert and accumulating high constraint violations due to failure to discriminate against unseen obstacle configurations absent from the expert demonstrations. MPAIL, as a planning-based method, shows greater robustness to out-of-distribution obstacle states represented as unseen obstacle occupancy than policy-based GAIL, but achieves higher cost rate than DBF-MPAIL. This is because DBF-based methods constrain the reward function search to satisfy the CBF dynamic constraint (\Cref{eq:dbf}), encouraging both policy and planning-based methods to learn safe imitative inhibitory behavior with minimal regret.

Notably, DBF-GAIL achieves the lowest cost rate at convergence, while DBF-MPAIL trades marginally higher cost rate for significantly better reward than DBF-GAIL. Although MPAIL achieves the highest reward, it does so at the cost of higher constraint violations compared to DBF-based methods, highlighting the safety-performance trade-off that DBFs are designed to address. AIRL and DBF-AIRL achieve comparable performance shown in \Cref{fig:airl_comparision}.

\textbf{Effect of different $\mathcal{K}$-class functions:}
AIL addresses the occupancy matching problem relative to the expert distribution, with the learned reward serving as a dynamic constraint for the Control Barrier Function (CBF). Since this reward directly affects safe exploration, we quantify its impact using the cost rate (regret) and total collisions per environment. 
In our experiments, discriminator logit outputs remained in the range $[-0.5, 0.0]$ across all considered $\mathcal{K}$-class functions (linear, cubic, and exponential) in training environment \Cref{fig:Q1_figure}b. Within this range, the slopes of these functions follow the order $\text{linear} > \text{cubic} > \text{exponential}$. As the reward is learned, the cost-function logits for generated trajectories shift from large negative values toward zero. This progression is reflected in the safety metrics, including cost rate and total collisions per environment, as shown in \Cref{fig:K-class}. The figure indicates that cost rate decreases fastest for linear and slowest for exponential, consistent with their slope ordering. While total collisions per environment continue to accumulate throughout learning, linear accumulates them at the lowest rate, further corroborating its superior safety performance. 
Together, these results demonstrate that DBF-based AIL methods learn to discriminate against unseen unsafe states, enabling safe online imitation learning whose inhibitory behavior is further modulated by the choice of $\mathcal{K}$-class function.

\subsection{Q3: Real World Validation Through Real-Sim-Real Experiment} \label{sec:ex_sub}
\begin{wrapfigure}{r}{0.65\columnwidth}
    \centering
    \includegraphics[width=\linewidth, frame]{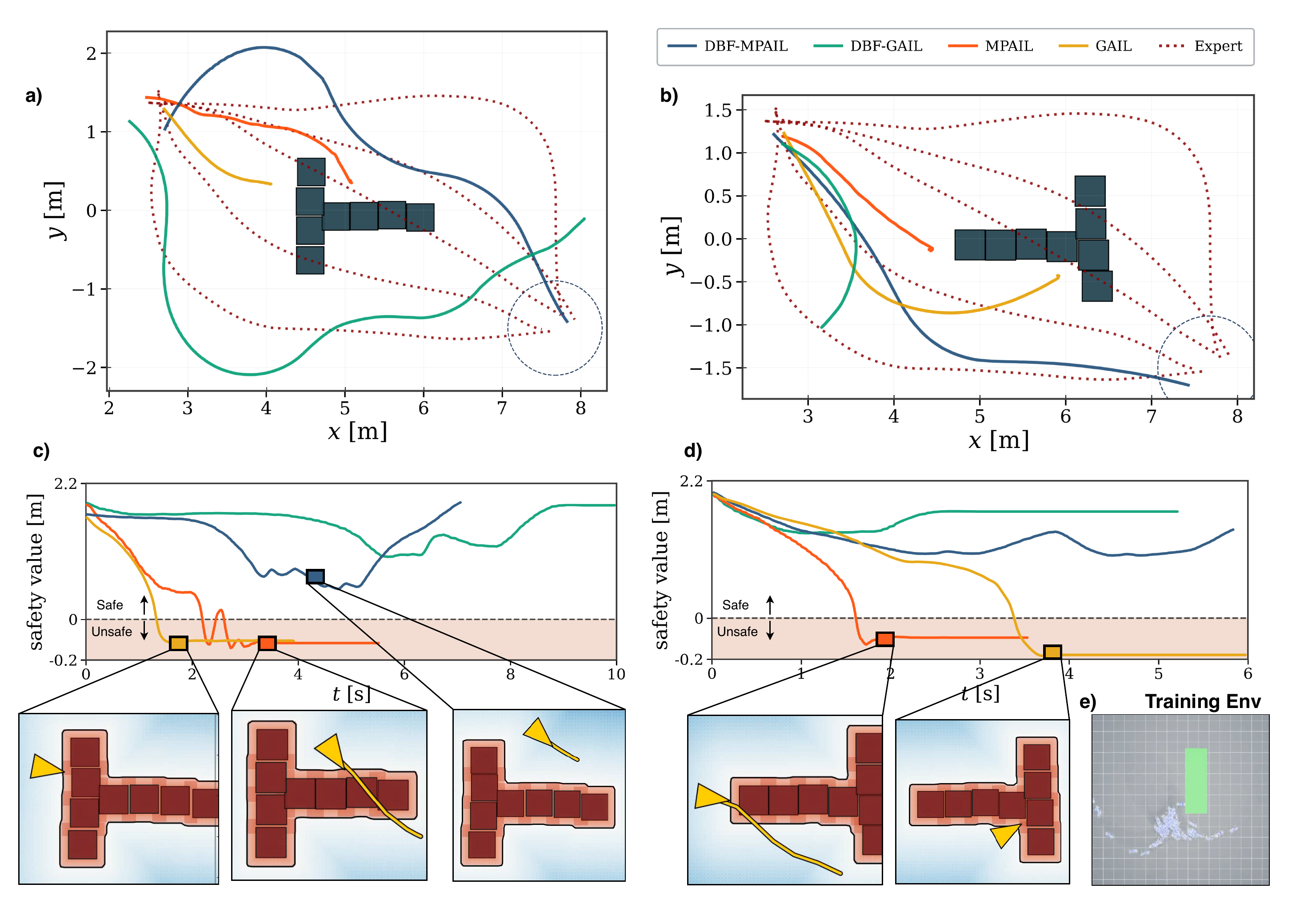}
    \caption{Hardware deployment across additional two obstacle  configurations than \Cref{fig:Hardware_validation}. (a,b) Closed-loop trajectories for all variants with expert reference (dotted). (c,d) Safety value represented by the minimum distance to all obstacles. Insets show robot state at near-collision timesteps. (e) Shows the training environment used to train all the methods.}
    \label{fig:real-sim-real}
\end{wrapfigure}
\textbf{Experiment Setup:} Our hardware experiment evaluates GAIL, MPAIL, DBF-GAIL, and DBF-MPAIL through a Real-to-Sim-to-Real pipeline: 1) a small number of state-only (position) observations are collected from the real world, 2) the method is trained using interactions from simulation with randomized obstacle positions, and 3) the method is deployed zero-shot in the real world for evaluation.

\textbf{Findings:}
We find that DBF-MPAIL consistently trace path during inference not directly present in the expert demonstrations, as shown in \Cref{fig:Hardware_validation}, to avoid obstacles while imitating the behavior needed to reach the goal. Unlike DBF-MPAIL, which plans over the barrier landscape, policy-based methods take positional state and a BEV map directly as input to a black-box policy.

\begin{wraptable}{r}{0.55\columnwidth}
\vspace{-12pt}
\centering
\scalebox{0.8}{
\begin{tabular}{lccc}
\toprule
\textbf{Variant} & \textbf{Succ.} $\uparrow$ & \textbf{Coll.}$\downarrow$ & \textbf{Min $d$ (m)} $\uparrow$ \\
\midrule
\grayrow
\multicolumn{4}{l}{\textit{Without DBF}} \\
GAIL      & 0/20~(0\%)   & 10/20~(50\%) & $0.035 \pm 0.274$\\
MPAIL     & 0/20~(0\%)   & ~6/20~(30\%) & $0.337 \pm 0.534$ \\
\midrule
\grayrow
\multicolumn{4}{l}{\textit{With DBF}} \\
DBF-GAIL  & ~1/20~(5\%)  & \textbf{~0/20~(0\%)}  & $0.500 \pm 0.437$ \\
DBF-MPAIL & \textbf{13/20~(65\%)} & \textbf{~0/20~(0\%)}  & $0.188 \pm 0.138$ \\
\bottomrule
\end{tabular}
}
\caption{Hardware deployment results (5 trials per variant across four obstacle configurations (\Cref{fig:Hardware_validation}, 
\Cref{fig:real-sim-real})). DBF variants achieve zero collisions across all trials. $d$ (m) denotes the mean minimum distance to obstacles throughout each trial}
\label{tab:hardware_results}
\vspace{-15pt}
\end{wraptable}
In addition, \Cref{fig:Hardware_validation} and \Cref{fig:real-sim-real} illustrate a key advantage of planning-based DBF-AIL: success in imitating goal-reaching behavior while avoiding obstacles unseen in the expert data and at unseen positions in the training environment. By granting access to the agent’s optimization landscape, DBF-MPAIL significantly improves the interpretability of agents trained on unlabeled and suboptimal data compared with black-box policies. We find that although the success rate of DBF-GAIL in reaching the goal is low due to the sim-to-real gap and the lack of structure in planning, it achieves zero collisions across all obstacle configurations, unlike the non-DBF baselines, as quantified in Table~\ref{tab:hardware_results}.
At deployment time, GAIL's policy frequently collides with obstacles or drifts away from the expert-demonstrated trajectory, undermining the imitation objective. Although real-world evaluations of AIL methods are scarce in the literature, our experiments indicate that AIL policies can degrade substantially when transferred to physical settings, consistent with the findings of \cite{sun_adversarial_2021}. As shown quantitatively in \Cref{tab:hardware_results}, DBF-based methods enhance the safety of imitation learning algorithms in real-world safety-critical deployments.



\section{Conclusion}
\label{sec:conclusion}


In this work, we introduce Discriminative Barrier Functions (DBFs), a framework for incorporating Control Barrier Function (CBF) theory directly into the AIL framework. By constraining the learned discriminator through barrier-function objectives, DBFs enable AIL methods to recover notions of safety and inhibitory control directly 
from unlabeled demonstrations without requiring explicit unsafe labels or hand-engineered safety costs. Across both policy-based and planning-based AIL algorithms, we demonstrate that DBFs improve the generalization of learned safe sets beyond the expert data distribution without sacrificing imitation performance. DBFs recover coherent safety boundaries, exhibit robustness to previously unseen obstacles and environmental changes, and induce safer exploration during online learning in both simulation and the real world. Overall, our results suggest that incorporating control-theoretic structure into adversarial imitation learning provides a promising direction toward safer and more robust online learning from observation.


\section{Limitations and Future Work}
\label{sec:limitations}



Although DBFs effectively learn to discriminate safe from unsafe states during learning and inference, the safety signal remains confined to the discriminator and does not directly shape the policy's value function. Future work will consider approximate Hamilton-Jacobi reachability methods~\citep{fixed_point_hjb_1, 
fixed_point_hjb_2} to modify the generalized advantage estimator, synthesizing a safety-aware value function for improved performance. Additionally, DBFs may become overly conservative when expert data covers only a narrow subset of valid strategies, incorrectly suppressing safe but underrepresented behaviors. Future 
work may investigate uncertainty-aware DBFs, active data collection, or mechanisms for reasoning about demonstration coverage and ambiguity. The $\mathcal{K}$-class function is currently specified by hand; prior work suggests these functions may themselves be learnable~\citep{learnable-k-class-fn}, potentially enabling more adaptive barrier representations across tasks with complex or heterogeneous safety constraints.


\clearpage


\bibliography{references}  

\newpage
\appendix
\section{Appendix}
\label{appendix:dbfs-derivation}

\paragraph{DBF as Cost-Regularized Observation-Only IRL.}

We now show how the DBF objective in \Cref{eq:dbf} can be interpreted as a restricted form of the entropy-regularized observation-only adversarial imitation learning formulation in \Cref{eq:irlfo}. 

To start, suppose that 

\begin{align}
    \mathcal{C}_\text{CBF} := \{ c : c(s, s') = q_h(s, s'), \; h \in \mathcal{H}\} \subset \mathbb{R}^{S \times S}
\end{align}

is a nonempty, closed, convex class of cost functions parametrized by $h$.

Now recall that Proposition 3.2 in \cite{ho_generative_2016} states that for any proper, closed, convex regularizer $\psi$, 

\begin{align}
    \text{RL} \circ \text{IRL}_\psi(\pi_E) = \arg\min_\pi [-H(\pi) + \psi^\ast(\rho_\pi - \rho_E)] 
    \label{eq:prop}
\end{align}

where $\text{IRL}_\psi(\pi_E) = \argmax_{c \in \mathbb{R^{S \times A}}} - \psi(c) + \min_{\pi \in \Pi} -\lambda \mathbb{H}(\pi) + \mathbb{E}_\pi [c(s, a)]- \mathbb{E}_{\pi_E}[c(s, a)]$ and $\rho_\pi$, $\rho_E$ are the state-action occupancy measures of the policy and expert respectively. Following \cite{torabi_generative_2019}, we instead consider the state-transition occupancy measure 

\begin{align}
    \rho_{\pi}(s, s') &= \sum_{a} P(s' \mid s, a)\,\pi(a \mid s) \sum_{t=0}^{\infty} \gamma^{t} P(s_t = s \mid \pi).
\end{align}

While \cite{torabi_generative_2019} derives the observation-only analogue of Equation~\ref{eq:prop} in the absence of entropy regularization, in practice we optimize the corresponding entropy-regularized objective

\begin{align}
    \text{RL} \circ \text{IRLfO}_\psi(\pi_E) = \arg\min_\pi [-H(\pi) + \psi^\ast(\rho_\pi - \rho_E)] 
    \label{eq:obs_prop}
\end{align}

where $\text{IRLfO}$ is defined in Equation~\ref{eq:irlfo} and $\rho_{\pi}$ and $\rho_E$ are now state-transition occupancy measures. This can be viewed as the natural entropy-regularized extension of the GAIfO objective.

Now consider the indicator cost function regularizer

\begin{align}
    \psi_\text{CBF}(c) &= \begin{cases}
        0, & \mathrm{if} \quad c \in \mathcal{C}_\text{CBF} \\
        \infty, & \mathrm{otherwise}
    \end{cases}
\end{align}

Note that this regularizer is proper, closed, and convex by assumption on $\mathcal{C}_\text{CBF}$. Now recall that 

\begin{align}
    \psi_\text{CBF}^\ast(u) &= \sup_{c \in \mathbb{R}^{S \times S}} [\langle u, c \rangle - \psi_{\text{CBF}}(c)] \\
\end{align}

But since $\psi_{\text{CBF}}(c) = 0$ for all $c \in \mathcal{C}_\text{CBF}$ and is $\infty$ everywhere else, we obtain that 

\begin{align}
    \psi_\text{CBF}^\ast(u) = \sup_{c \in \mathcal{C}_\text{CBF}} \langle u, c \rangle
\end{align}

Interpreting the discriminator as a parametrization of the cost function, we let $D_h := q_h$. Substituting $u = \rho_\pi - \rho_E$ and reparameterizing by $h \in \mathcal H$ yields

\begin{align}
    \psi_\text{CBF}^\ast(\rho_\pi - \rho_E) =  \sup_{h \in \mathcal{H}} [\mathbb{E}_{\rho_\pi}[D_h(s, s')] - \mathbb{E}_{\rho_E}[D_h(s, s')]]
\end{align}

From our earlier observation that $\psi_\text{CBF}$ is proper, closed and convex we may apply Equation~\ref{eq:obs_prop} and then obtain that 

\begin{align}
    \text{RL} \circ \text{IRLfO}_\psi(\pi_E) = \arg\min_\pi \max_{h \in \mathcal{H}} [-H(\pi) +  \mathbb{E}_{\rho_\pi}[D_h(s, s')] - \mathbb{E}_{\rho_E}[D_h(s, s')]] 
\end{align}

Furthermore, if we restrict the class of cost functions we consider to be 1-Lipschitz $\mathcal{C}_\text{CBF} \subset \{ c : \lVert c \rVert_L \leq 1  \}$ then the objective becomes an integral probability metric over the restricted CBF class. If additionally $\mathcal{C}_\text{CBF}$ equals the full 1-Lipschitz ball then by Kantorovich-Rubinstein duality we get that 

\[
    \sup_{c \in \mathcal{C}_\text{CBF} } [\mathbb{E}_{\rho_\pi}[c(s, s')] - \mathbb{E}_{\rho_E}[c(s, s')]] = W_1(\rho_\pi, \rho_E)
\]

where $W_1$ is the Wasserstein distance between the two distributions. Thus the discriminator objective corresponds to the Kantorovich dual formulation underlying WGANs.

\textbf{Remark:} In practice, we use the penalty term 

\begin{align}
    \lambda_{\text{GP}} \mathbb{E}_{\hat{\pi}} \Big[(\|\nabla_{(s, s')} D_{h}(s, s')\|_2 - 1)^2\Big]
\end{align}

to encourage approximate 1-Lipschitz continuity. Also, the exact parametrized neural network class used to represent $\mathcal{C}_\text{CBF}$ is not guaranteed to be convex or closed. However, the idealized function class can be approximated by:

\begin{align}
    \mathcal{H}_{\text{CBF}, \epsilon} &:= \{ h \in \mathbb{R}^S : h(s) \geq \epsilon \; \forall s \in \mathcal{S}, \; h(s) \leq -\epsilon \; \forall s \in \mathcal{U}\} \\
    \alpha(r) &= \kappa r + \beta \\
    \mathcal{Q}_\text{CBF} :&= \{ c : c(s, s') = q_h (s, s'), \; h \in \mathcal{H}_{\text{CBF}, \epsilon} \} \cap \{c : \lVert c \rVert_L \leq 1 \}
\end{align}

Then $\mathcal{H}_{\text{CBF}, \epsilon}$ is convex and closed as the intersection of half-spaces. It is also non-empty as long as $\mathcal{S}$ and $\mathcal{U}$ are disjoint and $\epsilon$ is small enough. Expanding the definition of $q_h$ yields $q_h(s, s') = h(s') + (\kappa -1 ) h(s) + \beta$ which is an affine map. We already know the Lipschitz ball is closed and convex and the parametrized cost function set is also closed and convex since $h \mapsto q_h$ is affine. Thus, $\mathcal{Q}_\text{CBF}$ is also closed and convex. Assuming there exists at least one feasible $h$ such that the induced $q_h$ is 1-Lipschitz, a mild assumption given the continuous setting of many robotics tasks, the class is nonempty so the assumptions on $\mathcal{C}_\text{CBF}$ are satisfied.

\section{Data Collection in Maze Environment}\label{appendix:Maze_env_data_collection}
\begin{wrapfigure}{r}{0.35\textwidth}
    \vspace{-50pt}
    \centering
    \includegraphics[width=\linewidth]{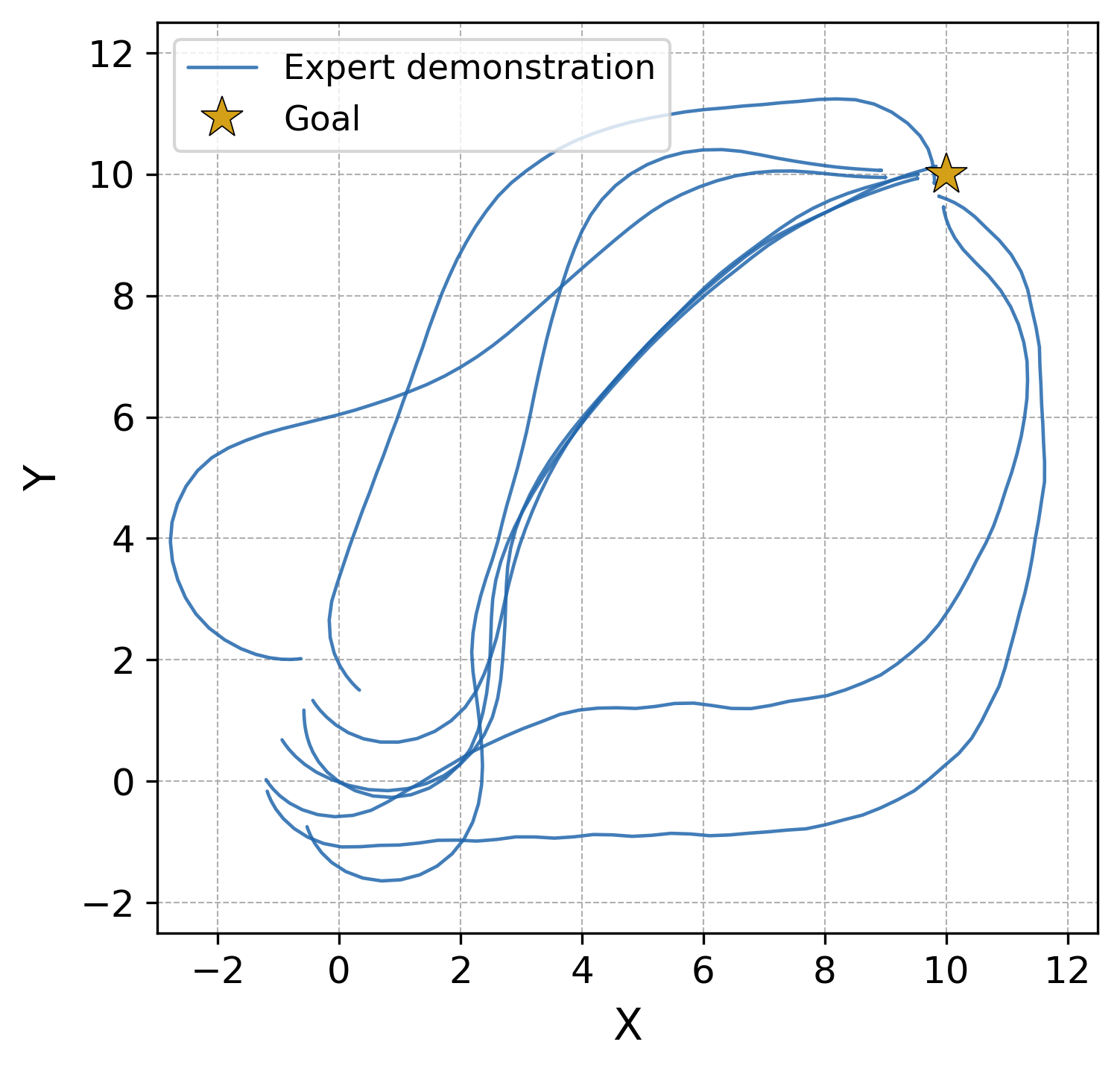}
    \caption{Expert trajectories collected in the maze environment by training PPO with varied wall configurations and reward heuristics.}
    \label{fig:expert_trajectory}
    \vspace{-60pt}
\end{wrapfigure}
To benchmark the AIL algorithms, we designed a maze environment and measured cumulative regret over the course of learning. Expert trajectories were collected by training PPO with a reward consisting of distance-to-goal and a collision penalty across diverse wall configurations. Using varied configurations produces a wider expert state-occupancy distribution, which is important for benchmarking: it allows both the policy and the planner to explore different trajectories when evaluated under randomized maze layouts. We save just state-only observations for training AIL algorithms.

\section{Comparison of AIRL and DBF-AIRL}
\begin{wrapfigure}{r}{0.5\textwidth}
    \vspace{-10pt}
    \centering
    \includegraphics[width=\linewidth]{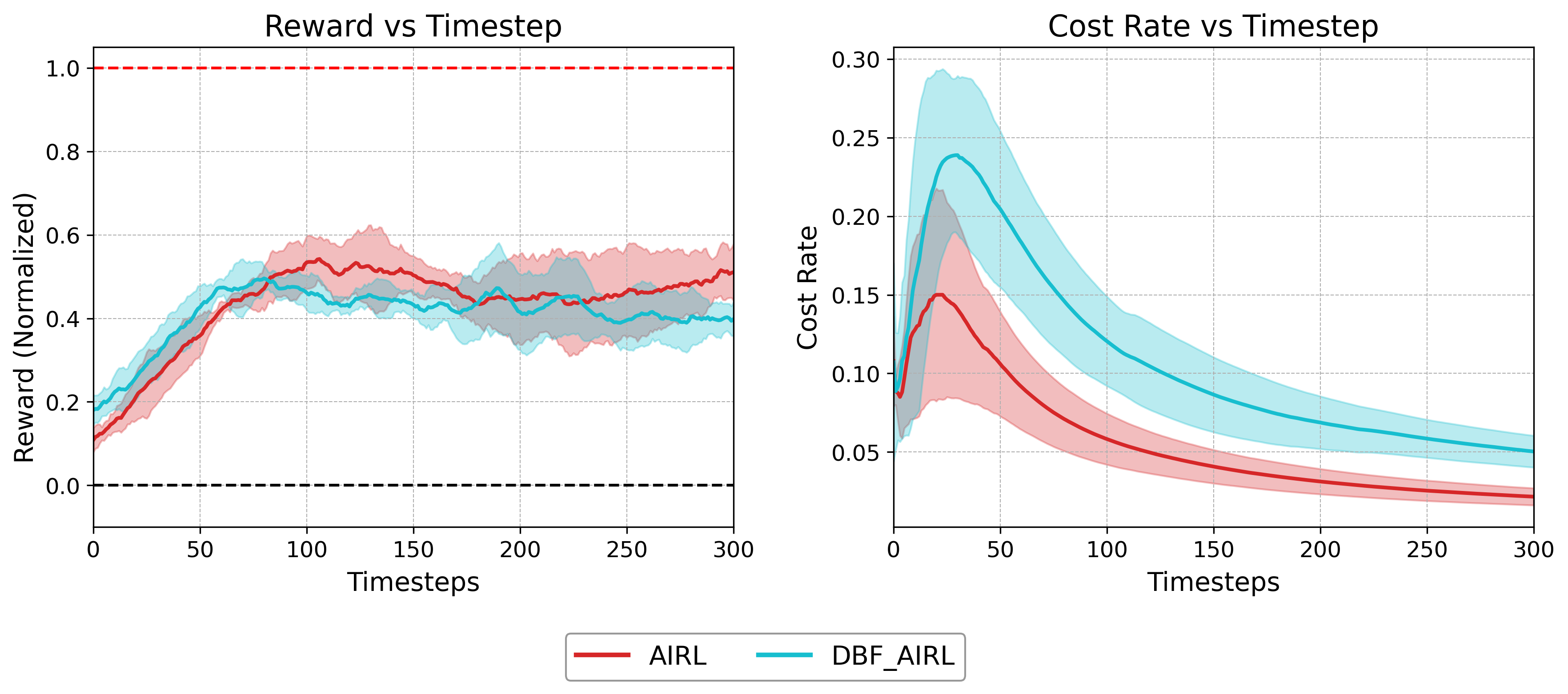}
    \caption{Performance comparison between AIRL and DBF-AIRL during online learning. AIRL achieves both higher reward and lower cost rate than DBF-AIRL, suggesting that AIRL's inherently conservative reward structure ($\sigma^{-1} \circ D$) already provides implicit safety regularization in this setting. Additionally, DBF-AIRL consistently decreases cost rate as learning progresses whilst maintaining comparable task performance.}
    \label{fig:airl_comparision}
    \vspace{-10pt}
\end{wrapfigure}
Both AIRL and DBF-AIRL achieve comparable performance during training (\Cref{fig:airl_comparision}) and evaluation (\Cref{tab:obstacle_eval}). Although DBF-AIRL incurs marginally higher regret during training, we hypothesize that safety context for black-box policies can be more robustly embedded via a safety value function based on the Hamilton-Jacobi fixed-point approximation \citep{paper:nakamura-cbf-2026, 
neural_cbf}, which we leave as future work. Furthermore, DBF-MPAIL and MPAIL share the same disentangled reward structure as AIRL -- important for sim-to-real transfer -- yet DBF-MPAIL clearly outperforms MPAIL in hardware evaluation (\Cref{tab:hardware_results}), supporting the claim that DBF yields a more safety-aware discriminative reward.

\section{Hyperparameters}\label{sec:hyperparams}
Hyperparameters for DBF-based methods in navigation task described in \Cref{fig:no-obstacle-sim-bf-comparison}. For baselines spectral normalization has been applied to maintain the Lipschitz continuity and is generally a common practice in making GAN or AIL training stable.
\begin{table}[h]
    \centering
    \scalebox{0.92}{
    \begin{tabular}{lccc}
    \toprule
      \textbf{Hyperparameter} & \textbf{GAIL / AIRL} & \textbf{DBF-GAIL / DBF-AIRL} \\
      \midrule
      Optimizer ($\theta$)      & Adam ($\beta_1{=}0.9$, $\beta_2{=}0.999$) & Adam ($\beta_1{=}0.9$, $\beta_2{=}0.999$) \\
      Learning rate             & $10^{-5}$ & $10^{-4}$ \\
      Hidden width              & 64  & 64 \\
      Hidden layers             & 3 & 3 \\
      Activation                & LeakyReLU  & SiLU \\
      Spectral norm             & On  & Off \\
      Disc.\ input              & GAIfO $(s, s')$ head & DBF $(s,s')$ head \\
      State slice               & xy  & xy\\
      Loss                      & BCE  & WGAN + ReLU spec + GP \\
      $\lambda_{\mathrm{WGAN}}$ & ---  & 1 \\
      $\lambda_{\mathrm{sign}}$ & --- & 5 \\
      $\lambda_{\mathrm{GP}}$   & ---  & 10 \\
      GP target                 & ---  & 1.0 \\
      $k-class$ function        & ---  & linear \\
      Mini-batches              & 3  & 3 \\
      Num epochs                & 100 & 100 \\
    \bottomrule
    \end{tabular}
    }
    \caption{Discriminator hyperparameters for simulated navigation task (GAIL, AIRL, DBF-GAIL/DBF-AIRL). DBF-GAIL and DBF-AIRL share identical discriminator training; only rollout rewards differ.}
\end{table}

The policy is trained with on-policy PPO. Actor and critic are updated jointly with an Adam optimizer; the learning rate is adaptively adjusted based on the approximate KL divergence between old and new policies ($\hat{D}_{\mathrm{KL}}$, target 0.01). 

\textbf{Policy inputs in the maze environment:}
In the obstacle avoidance task, the actor and critic receive asymmetric inputs. The ego vehicle has a BEV map centered on its position, whose obstacle occupancy updates dynamically with the environment. At each timestep, the actor receives the full 2-dimensional positional state vector concatenated with the flattened BEV map, while the critic receives the 2-dimensional positional state vector along with only the occupancy value of the center cell directly beneath the robot. This asymmetric design provides the critic with a local collision signal while allowing the actor to plan over the full spatial context.

\textbf{Discriminator and critic inputs in the maze environment:}
Across all experiments, whether policy-based or planning-based, the discriminator receives a $3$-dimensional input comprising the positional coordinates $x$, $y$, and the obstacle occupancy of the cell directly beneath the robot. The critic follows the same input structure across both policy-based and planning-based methods.

\begin{table}[h]
    \centering
    \scalebox{1.0}{
    \begin{tabular}{lc}
    \toprule
      \textbf{Hyperparameter} & \textbf{Value} \\
      \midrule
      Policy optimizer ($\pi$, $V$)   & Adam ($\beta_1{=}0.9$, $\beta_2{=}0.999$)\tablefootnote{Shared optimizer over actor and critic parameters (RSL-RL PPO).} \\
      Policy learning rate            & $10^{-3}$ (adaptive KL schedule) \\
      Desired KL                      & 0.01 \\
      Actor hidden layers             & [64, 64]\tablefootnote{Maze env: [1024, 512].} \\
      Critic hidden layers            & [64, 64]\tablefootnote{Maze env: [64, 64, 64].} \\
      Activation                      & ReLU \\
      Initial action std              & 1.0 \\
      PPO clip $\epsilon$             & 0.2 \\
      Value loss coefficient          & 1.0 \\
      Clipped value loss              & True \\
      Entropy coefficient             & 0.005 \\
      PPO mini-batches                & 4 \\
      PPO learning epochs             & 5 \\
      Discount $\gamma$               & 0.99 \\
      GAE $\lambda$                   & 0.95 \\
      Max gradient norm               & 1.0 \\
      Rollout length                  & 100 steps/env \\
    \bottomrule
    \end{tabular}
    }
    \caption{Policy (PPO) hyperparameters for GAIL, AIRL, DBF-GAIL and DBF-AIRL.}
\end{table}

\begin{table}[h]
    \centering
    \scalebox{0.92}{
    \begin{tabular}{lcc}
    \toprule
      \textbf{Hyperparameter} & \textbf{MPAIL} & \textbf{DBF-MPAIL} \\
      \midrule
      Optimizer ($\theta$)      & Adam ($\beta_1{=}0.5$, $\beta_2{=}0.999$) & Adam ($\beta_1{=}0.5$, $\beta_2{=}0.999$) \\
      Learning rate             & $10^{-5}$ & $5{\times}10^{-5}$ \\
      Hidden layers             & [32, 32] & [64, 64, 64] \\
      Activation                & LeakyReLU & SiLU \\
      Spectral norm             & On & Off \\
      DBF $\alpha$              & --- & linear \\
      Disc.\ input              & $(s, s')$ & $(s,s')$ \\
      State features            & 2-d (xy) & 2-d (xy) \\
      \midrule
      Disc.\ loss               & BCE & WGAN + ReLU spec + GP \\
      Mini-batches              & 3 & 3 \\
      Num epochs                & 100 & 100 \\
    \bottomrule
    \end{tabular}
    }
    \caption{Discriminator hyperparameters for simulated navigation experiment with MPAIL.}
\end{table}

\textbf{Planning Based AIL:} MPAIL is a recently introduced model-based AIL algorithm \citep{han_model_2025}. Algorithm 3 in the appendix of \citet{han_model_2025} describes the planning-based policy used in this work. For DBF-MPAIL, the only changes are the reward function (i.e.\ the discriminator), which encodes a CBF-based safety constraint, and the use of an off-policy replay buffer in place of MPAIL's on-policy buffer.

\begin{table}[h]
    \caption{MPAIL vs.\ DBF-MPAIL hyperparameters for obstacle (maze) navigation.}
    \centering
    \scalebox{0.88}{
    \begin{tabular}{lcc}
    \toprule
      \textbf{Hyperparameter} & \textbf{MPAIL} & \textbf{DBF-MPAIL} \\
      \midrule
      \multicolumn{3}{l}{\textit{Training}} \\
      Rollout length              & 100 & 100 \\
      Training iterations         & 300 & 300 \\
      Replay buffer               & --- & 200{,}000 transitions \\
      Replay batch size           & --- & 32 \\
      \midrule
      \multicolumn{3}{l}{\textit{Discriminator}} \\
      Optimizer ($\theta$)        & Adam ($\beta_1{=}0.5$, $\beta_2{=}0.999$) & Adam ($\beta_1{=}0.5$, $\beta_2{=}0.9$) \\
      Learning rate               & $10^{-5}$ & $7{\times}10^{-5}$ \\
      Architecture                & GAIfO MLP & DBF cost net \\
      Hidden layers               & [64, 64, 64] & [64, 64, 64] \\
      Activation                  & LeakyReLU & SiLU \\
      Spectral norm               & On & Off \\
      State features              & xy + map (2-d) & xy + map (2-d)\tablefootnote{Map is a 2D bird's-eye-view occupancy grid of size $4.1\,\text{m} \times 4.1\,\text{m}$; the discriminator receives the occupancy value of the cell beneath the robot as a scalar input.} \\
      DBF $\alpha$                & --- & linear \\
      Loss                        & BCE & WGAN + ReLU spec + GP \\
      Mini-batches / epochs       & 3 / 100 & 3 / 100 \\
      $\lambda_{\mathrm{WGAN}}$   & --- & 1 \\
      $\lambda_{\mathrm{sign}}$   & --- & 5 \\
      $\lambda_{\mathrm{GP}}$     & --- & 10 \\
      GP target                   & --- & 1.0 \\
      \midrule
      \multicolumn{3}{l}{\textit{Value function}} \\
      Optimizer / lr              & Adam, $5{\times}10^{-3}$ & Adam, $5{\times}10^{-3}$ \\
      Network                     & [24, 24, 24], ReLU & [24, 24, 24], LeakyReLU \\
      Value clip                  & 0.2 & 0.2 \\
      Discount $\gamma$           & 0.99 & 0.99 \\
      GAE $\lambda$               & 0.95 & 0.95 \\
      \midrule
      \multicolumn{3}{l}{\textit{MPPI policy}} \\
      Rollouts / horizon          & 512 / 11 & 512 / 11 \\
      Temperature / opt.\ iters.  & 0.1 / 5 & 0.1 / 5 \\
      Action std (init / min)     & 1.0 / 0.001 & 1.0 / 0.001 \\
      Sampling noise              & [0.5, 0.5] & [0.5, 0.5] \\
      Control limits $(v,\omega)$ & $(0, 1)$, $(-1, 1)$ & $(0, 1)$, $(-1, 1)$ \\
      BEV map size                & $4.1m \times 4.1m$ & $4.1m \times 4.1m$ \\
    \bottomrule
    \end{tabular}
    }
\end{table}

\section{Hardware Experiment Details}
Our hardware experiments are conducted in a motion capture room, which provides ground-truth position and obstacle locations in the environment. We use the open-source $1/10$-scale race car MUSHR \citep{srinivasa2019mushr}, with its onboard compute replaced by an NVIDIA Jetson Orin NX. The platform uses a bicycle model for forward rollouts with a 12-dimensional state (position, orientation, linear and angular velocities), and a 2-dimensional positional state $(x, y)$ with obstacle occupancy as input to the discriminator (reward function or MPPI stage cost) and value function (in model-based AIL). The vehicle is operated without slipping or reversing, and only positional data is collected as the expert dataset $\mathcal{D}_E = \{(x_t, y_t)\}$. All state information from the motion capture system is received at 20\,Hz, and the bird's-eye-view (BEV) occupancy map is updated at the same frequency by querying obstacle positions within the map region.

\begin{figure}
    \centering
    \includegraphics[width=\textwidth]{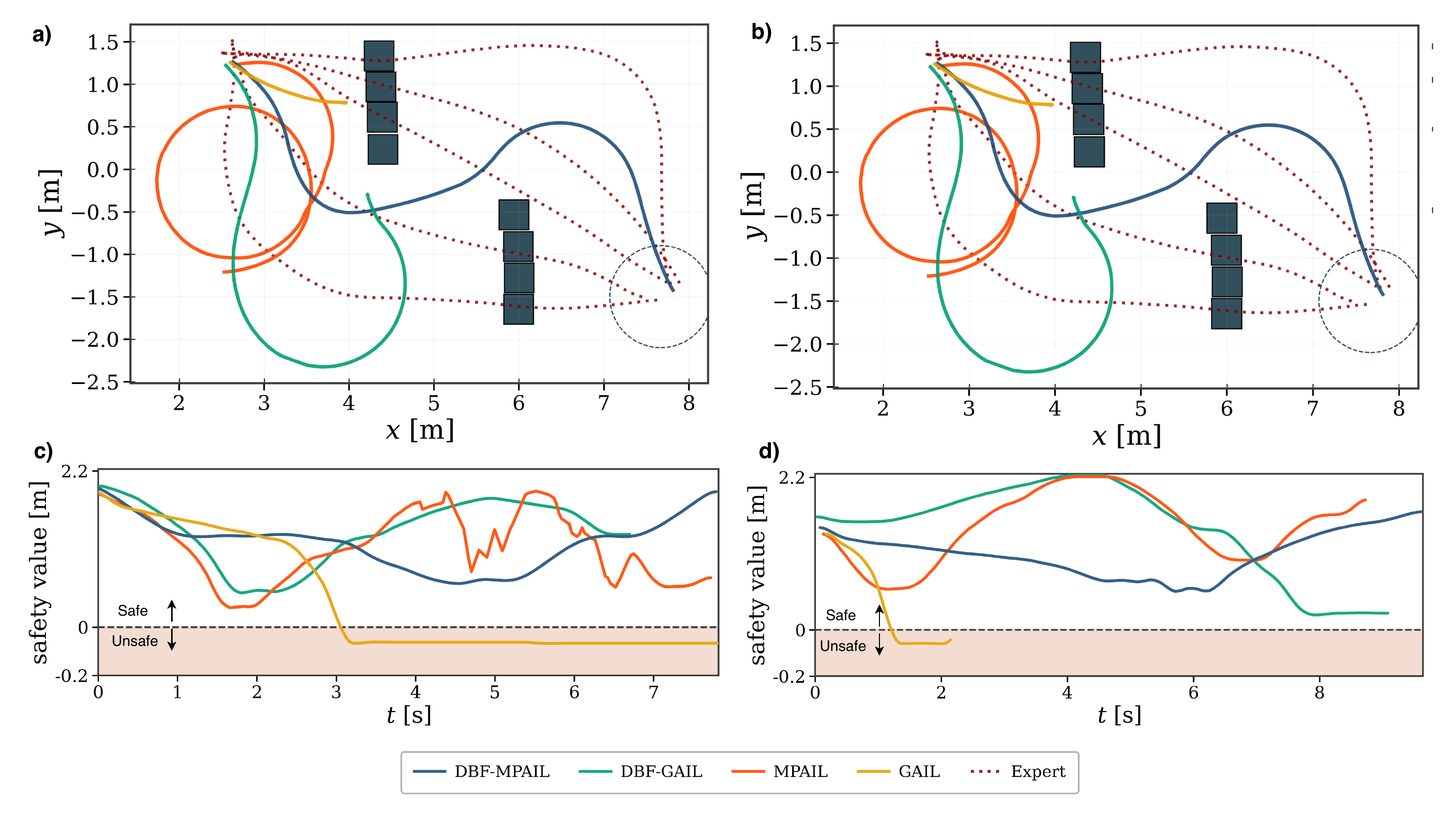}
    \caption{Extension of \Cref{fig:real-sim-real} illustrating baseline trajectories across two hardware scenarios. \textbf{(a,b)} Closed-loop rollouts of all methods in scenarios 1 and 2, respectively. \textbf{(c,d)} Minimum distance to obstacles over time for scenarios 1 and 2, respectively. DBF-MPAIL successfully navigates both scenarios, demonstrating generalization to unseen obstacle configurations. DBF-GAIL avoids collisions but fails to reach the goal due to the sim-to-real gap. MPAIL and GAIL fail to imitate the expert trajectory in both scenarios. \textbf{Note:} Expert demonstrations were collected in an obstacle-free environment.
\label{fig:baselines_plots}}
\end{figure}
\end{document}